\documentclass[nohyperref]{article}

\usepackage{microtype}
\usepackage{graphicx}
\usepackage{subfigure}
\usepackage{booktabs} 

\usepackage{hyperref}
\usepackage{adjustbox}

\usepackage[accepted]{icml2022}

\usepackage{amsmath}
\usepackage{amssymb}
\usepackage{mathtools}
\usepackage{amsthm}
\usepackage{multirow}
\usepackage[capitalize,noabbrev]{cleveref}

\theoremstyle{plain}

\theoremstyle{definition}

\theoremstyle{remark}

\usepackage[textsize=tiny]{todonotes}

\icmltitlerunning{DR-Label: Improving GNN Models for Catalysis Systems by Label Deconstruction and Reconstruction}

\begin{document}

\twocolumn[
\icmltitle{DR-Label: Improving GNN Models for Catalysis Systems \\ by Label Deconstruction and Reconstruction}






\icmlsetsymbol{equal}{*}

\begin{icmlauthorlist}
\icmlauthor{Bowen Wang}{equal,cuhk}
\icmlauthor{Chen Liang}{equal,tbsi}
\icmlauthor{Jiaze Wang}{cuhk}
\icmlauthor{Furui Liu}{zjlab}
\icmlauthor{Shaogang Hao}{tencent}
\icmlauthor{Dong Li}{huawei}
\icmlauthor{Jianye Hao}{huawei}
\icmlauthor{Guangyong Chen}{zjlab}
\icmlauthor{Xiaolong Zou }{tbsi}
\icmlauthor{Pheng-Ann Heng}{cuhk,zjlab}
\end{icmlauthorlist}

\icmlaffiliation{cuhk}{Department of Computer Science and Engineering, The Chinese University of Hong Kong, Hong Kong, China}
\icmlaffiliation{tbsi}{Shenzhen Geim Graphene Center, Tsinghua-Berkeley Shenzhen Institute \& Tsinghua Shenzhen International Graduate School, Tsinghua University, Shenzhen, China}
\icmlaffiliation{huawei}{Huawei Noah's Ark Lab, Shenzhen, China}
\icmlaffiliation{zjlab}{Zhejiang Lab, Hangzhou, China}
\icmlaffiliation{tencent}{Tencent, Shenzhen, China}

\icmlcorrespondingauthor{Guangyong Chen}{gychen@zhejianglab.com}

\icmlkeywords{Machine Learning, ICML}

\vskip 0.3in
]




\printAffiliationsAndNotice{\icmlEqualContribution} 

\begin{abstract}

Attaining the equilibrium state of a catalyst-adsorbate system is key to fundamentally assessing its effective properties, such as adsorption energy. Machine learning methods with finer supervision strategies have been applied to boost and guide the relaxation process of an atomic system and better predict its properties at the equilibrium state. In this paper, we present a novel graph neural network (GNN) supervision and prediction strategy \textbf{DR-Label}. The method enhances the supervision signal, reduces the multiplicity of solutions in edge representation, and encourages the model to provide node predictions that are graph structural variation robust. DR-Label first \textbf{D}econstructs finer-grained equilibrium state information to the model by projecting the node-level supervision signal to each edge. Reversely, the model \textbf{R}econstructs a more robust equilibrium state prediction by transforming edge-level predictions to node-level with a sphere-fitting algorithm. The DR-Label strategy was applied to three radically distinct models, each of which displayed consistent performance enhancements. Based on the DR-Label strategy, we further proposed DRFormer, which achieved a new state-of-the-art performance on the Open Catalyst 2020 (\textbf{OC20}) dataset and the Cu-based single-atom-alloyed CO adsorption (\textbf{SAA}) dataset. We expect that our work will highlight crucial steps for the development of a more accurate model in equilibrium state property prediction of a catalysis system.

\end{abstract}
\section{Introduction}

\begin{figure*}[ht!]
\begin{center}
\includegraphics[width=\textwidth]{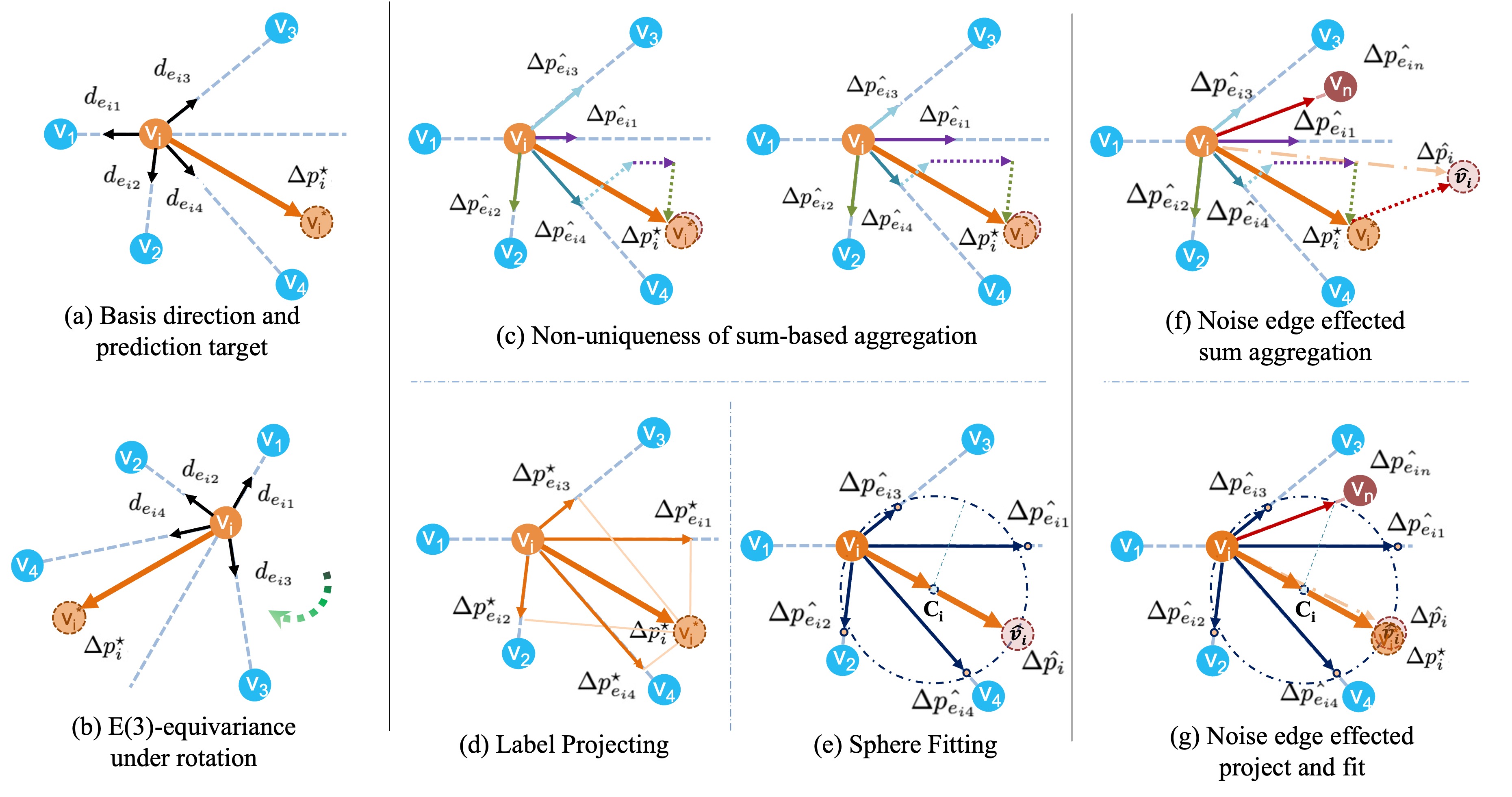}
  \caption{The general idea of DR-Label. (a) and (b), illustration of the required E(3)-Equivariance for N-body state prediction models. (c) Illustration of sum-aggregated node prediction, and the non-uniqueness of edge representation. (d) and (e), the idea of DR-Label strategy.  (f) and (g), the result comparison of sum-based aggregation and DR-Label under graph structural variation.
  }
  \label{pnf idea}
\end{center}
\end{figure*}


 Discovering and developing new and efficient catalysts is an urgent topic nowadays\cite{newell2020global_catalyst_urgent}, while numerous amounts of resources are required to achieve that target by chemical synthesis. First-principles calculations are applied widely in that area to compute catalytic performances of materials before they are prepared, so that experimentalists can focus on only the high-performance ones. However, the prohibitively high computational costs of first-principles calculations methods, such as density functional theory (DFT), make it impractical to explore the whole configuration space\cite{matera2019progress_catalyst_simulation}. In that case, machine learning has now become a promising solution to discovering potential catalysts for its fast inference ability based on the accumulated databases.

The key to catalyst effectiveness assessment is to attain its adsorption energy towards specific adsorbate, which is a property in the equilibrium state of the catalyst-adsorbate system\cite{chanussot2021OC20}. The adsorption energy $E_{ads}$ is defined by $E_{ads}=E^\star_{comb}-E^\star_{mol}-E^\star_{slab}$: the energy of combined catalyst slab and adsorbate compounds at the relaxed state $E^\star_{comb}$, minus the isolated relaxed energy of adsorbate $E^\star_{mol}$ and relaxed slab $E^\star_{slab}$. During the structure relaxation procedure of DFT calculations, an initial state of the system $\mathcal{G}$, in which the adsorbate is placed on the ideal catalyst surface, is provided as input. Positions of atoms in the system are then updated iteratively based on a gradient descent algorithm to search for an equilibrium (relaxed) state $\mathcal{G}^\star$, where the energy of the system is the lowest and forces experienced by atoms are balanced.

Graph neural network (GNN) based models currently play a preponderant role in equilibrium state property predictions. Early efforts modeled the atomic system with a graph structure and directly predict the target system-level energy\cite{MPNN,xie2018CGCNN}. It is a tendency that researchers tend to push a more elaborated message passing and representations for models, such as involving more steps of neighborhood information in each message-passing step \cite{schutt2017schnet,klicpera2020dimenet,klicpera2021gemnet}, or using more advanced representations in more detailed graph components \cite{TFN, SCN, GVP}. Moreover, it is shown in previous literature that jointly adding the supervision signal of the geometry information at the equilibrium state can largely boost the performance of the equilibrium property prediction. Many works such as \cite{noisynodes, equiformer, EGNN} added atomic-level supervision signals towards the equilibrium positions, while other methods extended the supervision signals to the inter-atomic level \cite{mtmd-saa, mendez2021geometricDeepDock}. However, for models extended to inter-atomic supervision, the edge-level predictions and node-level predictions are separately supervised by different heads, without underlying coupled mechanisms. For models with atomic-level supervisions, existing GNN methods commonly deploy linearly aggregated node predictions from edges, where representation ambiguity can be largely induced. More importantly, the graph representation of the same atomic system largely varies under different algorithms or different settings\cite{gemnet-oc,graphormer,MPNN}. The non-unique nature of graph representation for atomic systems further enhances the difficulty of obtaining an idea edge representation, making the atomic positional predictions largely volatile under graph structural change.

In order to develop a better-performed GNN model for catalyst system equilibrium state prediction, in this work we study the principles of providing an E(3)-equivariant label deconstruction strategy that expands the atomic-wise equilibrium geometric state signal to the inter-atomic level, thereby bringing a finer-grained supervision signal to models. During inference, we first generate the inter-atomic level prediction, then reconstruct the node-level prediction through an inverse operation of the deconstruction process. Our deconstruction and reconstruction (\textbf{DR-Label}) strategy avoids the multiplicity of solutions in the existing linear aggregated node prediction while being more robust under graph structural variation, largely improving the supervision towards the equilibrium state. We examined our methods using the Open Catalyst 2020 (OC20) \cite{chanussot2021OC20} dataset and the Cu-based single-atom-alloyed CO adsorption (SAA) dataset \cite{mtmd-saa} in simulating the relaxation and adsorption energy prediction process of a catalyst system. We provided a simple-to-implement instantiation of the DR-Label strategy and integrated it into three radically different cutting-edge models. Each model exhibited a consistent performance enhancement by applying our method. Utilizing the graph structure robustness property of DR-Label, we further developed DRFormer, which achieved a new state-of-the-art performance on OC20 and SAA datasets. These results suggest that DR-Label can serve as a crucial component for catalyst adsorption energy prediction models, with the potential of further generally applied to other state-property prediction tasks of N-body systems.

\section{Notation and Preliminaries}







In this section, we introduce the notations and preliminaries for the problem of equilibrium state prediction of an atomic system.


Let $\mathcal{G} =(\mathcal{V}, \mathcal{E})$ be the graph representation of an N-body atomic system with $N$ atoms, where each element in the node set $\mathcal{V} = \{v_1, v_2, ..., v_N\}$ represents each atom. 
$v_i$ contains two part of feature$\{h_i,p_i\}$: $h_i\in\mathbb{R}^d$ refers the E(3)-invariant atom properties, and $p_i\in\mathbb{R}^3$ is the atomic position in the Euclidean space. $\mathcal{E} = \{e_1, e_2, ..., e_{|\mathcal{E}|}\}$ is the set of directed edges between nodes, where for each edge $e_{ij}=(v_i, v_j)$, $v_i, v_j \in \mathcal{V}$ are the sender node and receiver node, respectively, and $\mathbf{m}_{ij}$ is the corresponding edge embedding. $\mathcal{N}_i$ refers to the set of 1-step neighbor nodes of $v_i$. 



For an N-body atomic system, the set of atom positions $\mathcal{P}^{0}=\{p_1^0, p_2^0, ..., p_N^0\}$ are dynamically changed to the equilibrium state $\mathcal{P^\star}=\{p_1^\star, p_2^\star, ..., p_N^\star\}$ due to the inter-atomic forces in structure relaxation, and $\Delta\mathcal{P}^\star=\{\Delta p_i^\star = p_i^\star-p^0_i|i\in 1,2,...,N\}$. In our task, machine learning models are applied to intake $\mathcal{G}^0$ and generate a graph-level output $\hat{\mathbb{G}}$ to minimize $||\mathbb{G}^\star-\hat{\mathbb{G}}||$, where $\mathbb{G}^\star$ is the corresponding graph-level property at the equilibrium state $\mathcal{G}^\star$. A geometric state forecast branch is commonly deployed to help the model to learn the underlying physical rules, where the model produces a new set of atomic positions $\mathcal{\hat{P}}=\{\hat{p_1}, \hat{p_2}, ..., \hat{p_N}\}$ that minimizes $\sum_{i\in N}{||p^\star_i-\hat{p_i}||_2}$. Given the initial state, the problem of the equilibrium state prediction of an atomic system is to develop a model that simultaneously predicts the system-level property $\mathbb{G}^\star$ and atomic positions $\mathcal{P}^\star$.


\section{Related Work}
\label{sec_related}

\subsection{Catalyst Adsorption Energy Prediction}


Early exploration of machine learning-based material property prediction such as CGCNN \cite{xie2018CGCNN} focuses on bulk materials and constructs the graph that incorporates periodic boundary conditions in a topological perspective. MEGNet \cite{chen2019MEGNet} proposed a universal framework for molecules and crystals that update edge, node, and graph level representations within the model. These methods heavily rely on expert-engineered features, which are found to be non-essential in recent studies \cite{klicpera2020dimenet}. Currently, popular catalyst adsorption energy prediction methods include the iterative relaxation method and the direct method. 

The iterative relaxation method, such as \cite{SCN,klicpera2021gemnet,gemnet-xl,gemnet-oc}, trains models as an imitation of DFT calculation. It predicts the transient energy and force of the atomic system, and iteratively updates atomic positions to the relaxed state. Though more accurate, the computational cost for training and inference can be prohibitively large, where SCN requires up to 1280 GPU days for training while GemNet-XL requires 1962 GPU days. On the contrary, the direct method intakes the initial state and directly predicts the relaxed energy, reducing the training cost to tens of GPU days. Based on \cite{graphormer}, \cite{graphormer-3d} added node-level supervision to explicitly learn the relaxed state, and won the championship of OC20 contest \cite{chanussot2021OC20}.  \cite{noisynodes} added a noise-correcting node-level loss as an over-smoothing proof regularizer, which essentially boosted the performance of \cite{gns} on adsorption energy prediction models. \cite{equiformer} further achieved superior performance by incorporating equivariant features and novel equivariant attention.

It is notable that for direct adsorption energy prediction, geometric forecast of relaxed state $\mathcal{G}^\star_{comb}$ plays an especially crucial role. This is because the adsorption energy of a catalyst system is computed based on the relaxed state of the system $E_{ads}=READOUT(\mathcal{G}^\star_{comb})$. \cite{equiformer}\cite{noisynodes} revealed that given the initial state $\mathcal{G}^0_{comb}$, auxiliary supervision of node positional displacement $\Delta\mathcal{P}^\star$ towards the relaxed state largely improved the model performance. \cite{mtmd-saa} showed that compared with using an initial state as input, using the relaxed state could provide up to 40\% decrease of MAE on adsorption energy prediction under the same model. 

\subsection{Invariant and Equivariant GNNs}

Developing an effective machine learning-based model to predict the equilibrium state and property is of high interest far beyond the catalyst adsorption energy prediction area. Similar models and techniques can also be transferred to other pivotal topics such as drug discovery \cite{Mendez-Lucio2021deepdock,uni-mol}, protein structure forecast \cite{GVP,Jumper2021alphafold}, and even in engineering design \cite{pfaff2020meshgraphnet}. 
Graph neural networks (GNNs) currently play a predominant role in this area. For these problems,  group invariance and equivariance are crucial properties that must be carefully considered during model development. An E(3)-equivairant model $\mathbf{f_{E(3)}}$ should fulfill the condition that for a E(3) group operation $\psi$, we have $\mathbf{f_{E(3)}}(\psi(\mathcal{P}))=\psi(\mathbf{f_{E(3)}}(\mathcal{P}))$. 

Due to their lack of group equivariance, early models such as \cite{gilmer2017MPNN} cannot be used for geometric state forecasting. Existing top-performed models preserve E(3) equivariance in two main directions. The first focuses on proposing equivariant models based on message-passing strategies. SchNet \cite{schutt2017schnet} started from Schrodinger's equation and used continuous edge encoders to ensure the solution space lies within real-world physical space. DimeNet \cite{klicpera2020dimenet} further encoded both the inter-atom distance and angular information through message passing and achieved comparable or higher performance of DFT calculations on QM9 dataset experiments. EGNN \cite{EGNN} linearly aggregated edge representation with edge directions to explicitly update the geometric feature of each node and achieved E(n)-equivariance. More recently, based on DimeNet++ \cite{klicpera2020dimenetpp}, GemNet \cite{klicpera2021gemnet} further added dihedral angle information in representation update and attained superior performances. Another direction focuses on finer equivariant representation designing \cite{equiformer} \cite{TFN}; \cite{nequip}; \cite{se(3)-transformers}. Spherical harmonics are commonly used in these methods for its equivariance property. \cite{GVP} additionally added the vector channel besides the scalar representation for each node. More recently, \cite{SCN} proposed to represent each node as a $(L+1)^2 \times C$ matrix, which represents $C$ channels of functions on the unit sphere, while each sphere function is parameterized by $L$ degree of spherical harmonic basis.

\section{Method}
\label{method}

In this section, we stress the limitations of existing models that generate linear aggregated geometric state forecasts. Then we propose principles of performing label deconstruction and reconstruction (DR-Label) that improve GNN models. Finally, we instantiate an algorithm named project and fit that implements the DR-Label principle to improve the GNN models for catalysis systems. 

\subsection{Limitation of Linear Aggregated Node Prediction}

As illustrated by Figure \ref{pnf idea} (a) and (b), in order to generate an E(3)-equivariant node-level positional displacement prediction, we summarize the general framework of most existing methods as follows:

\begin{equation}
\mathbf{m}_{ij} =\mathbf{f}_{emb} \left( \mathcal{G} \right) \\,
\end{equation}

\begin{equation}
\mathbf{\overrightarrow{m}}_{ij} =\mathbf{f}_{vec}\left(
\mathbf{f}_{geom}\left(p^0_i,p^0_j\right),
\mathbf{m}_{ij}\right) \\,
\end{equation}

\begin{equation}
\label{equation3}
\hat{p}_i =p_i^0+
\mathbf{f}_{pos}\left( \sum_{j \neq i} \mathbf{\overrightarrow{m}}_{ij}\right) \\.
\end{equation}



 The general idea is to treat the incident outward-edge directions $d_{e_{ij}}=(p_i-p_j)/|p_i-p_j|$ of node $v_i$ as the local frame because the edge directions are E(3)-equivariant. Existing methods first generate edge embedding $\textbf{m}_{ij}$ for each edge based on the initial graph representation of the system. Then, an edge-direction encoding $\mathbf{f}_{geom}(\mathbf{p}_i,\mathbf{p}_j)$ will combine the edge encoding $\textbf{m}_{ij}$, and feed into the directional edge embedding function $\mathbf{f}_{vec}$ to further ensure the final geometric information is encoded in the $\mathbf{\overrightarrow{m}}_{ij}$. Finally, the existing method performs a sum aggregation over each incident edge and transforms the result by $\mathbf{f}_{pos}$ to generate a positional displacement vector $\Delta \hat{p}_i=\mathbf{f}_{pos}\left( \sum_{j \neq i} \mathbf{\overrightarrow{m}}_{ij}\right)$

This pipeline provides a straightforward local-frame-based approach for generating node predictions that are E(3)-equivariant. However, the vanilla supervision signals of the linear aggregated method for geometry state prediction lie within the scope of the node level. Therefore, as illustrated in Figure \ref{pnf idea}(c), the physical significance of inter-atomic embeddings is unclear, leading to multiple interpretations of edge-level embeddings, i.e., there exists more than one set of edge representations ${m_{ij}, j\in\mathcal{N}_i}$ that can generate the same node representation. This property is problematic because the graph representation of an atomic system is naturally non-unique, where the variation of graph structure is inevitable and ubiquitous.

For instance, different graph construction methods can result in diverse graph structures. Most methods construct the graph based on the Euclidean distance between atoms, i.e., edge $e_{ij}$ is introduced wherever $|p_i-p_j|$ is smaller than a cutoff distance $C \in \mathbb{R}$\cite{klicpera2020dimenet,klicpera2020dimenetpp,klicpera2021gemnet}, while GemNet-OC proposes constructing graphs with a fixed number of nearest neighbors (e.g., GemNet-OC\cite{gemnet-oc}) to ensure a more consistent graph statistics across instances. On the other hand, \cite{graphormer} constructs a nearly fully connected graph. Moreover, the varied geometric statistics of different instances can also bring large variations to the topological statistics of the graph structure, making some graphs too dense while others are disconnected. For fully connected graphs, the average node degree will be directly influenced by the number of atoms in the system. Even when the same graph construction algorithm is deployed, the geometric structure iteration process during relaxation can also induce varied inter-atomic distances, leading to abundant edge addition or dropping.

Under these circumstances, the optimal model would be difficult to obtain since the ideal edge prediction is not unique. Furthermore, linear aggregated node predictions are highly sensitive to the ever-changing graph structure. As shown in Figure \ref{pnf idea} (f), suppose a new edge $e_{in}=(v_i, v_n)$ is induced to $v_i$ due to a change of the graph structure, which generates an edge-wise prediction $\Delta \hat{p}{e{in}}$. For a sum-aggregated node displacement prediction, the deviation from the original prediction will be equal to $\Delta \hat{p}_{e_{in}}$. Hence, it is imperative to provide the model with clear and unique edge-wise supervision that can withstand structural changes in the graph. Conversely, we expect that the node predictions aggregated from edges will also be robust to variations in the graph structure.

\subsection{Principles of Label Deconstruction and Reconstruction}
\begin{algorithm}[t]
   \caption{Deconstruction: Label Projecting ($\phi$)}
   \label{alg:proj}
\begin{algorithmic}
   \STATE {\bfseries Input:} $\mathcal{P}^0, \Delta\mathcal{P}^\star\in \mathbb{R}^{N\times 3}$ - the initial position and positioanl shift of each atom, $\mathcal{E}$ - directed edges between graph nodes.
        \FOR{$i$ in $N$}
        \FOR{$j$ in $\mathcal{N}_i$}
        \STATE $d_{e_{ij}}=(p^0_i - p^0_j)/|p^0_i - p^0_j|$
        \STATE $\Delta p^\star_{e_{ij}} = \Delta p^\star_i \cdot d_{e_{ij}} \cdot d_{e_{ij}}$
        \ENDFOR
        \ENDFOR
        \STATE \textbf{Return} $\{\Delta p^\star_{e_{ij}} \in \mathbb{R}^3: e_{ij}\in\mathcal{E}\}$ - The set of projected vectors from nodes to edges
\end{algorithmic}
\end{algorithm}

\begin{algorithm}[t]
   \caption{Reconstruction: Sphere Fitting ($\phi^{-1}$)}
   \label{alg:fit}
\begin{algorithmic}
   \STATE {\bfseries Input:} $\{\Delta \hat{p_{e_{ij}}} \in \mathbb{R}^3: e_{ij}\in\mathcal{E}\}$ - prediction of positional shift projection on edges, $\mathcal{P}^0 $ - initial position.
        \FOR{$i$ in $N$}
        \STATE Create $b=\mathbf{0}\in \mathbb{R}^3$, $ A=\mathbf{0}\in \mathbb{R}^{3 \times 3}$
        
        \FOR {$j$ in $|\mathcal{N}_i|$}
        \STATE $v = v+ (\Delta \hat{p_{e_{ij}}}^\top \Delta \hat{p_{e_{ij}}}) \cdot \Delta \hat{p_{e_{ij}}} / |\mathcal{N}_i|$
        \STATE $A = A+ \Delta \hat{p_{e_{ij}}} \Delta \hat{p_{e_{ij}}}^\top / |\mathcal{N}_i|$

        \ENDFOR
        
        
        \STATE $C_i = (2\cdot A)^{-1} \cdot b$
        \STATE $\hat{p_{i}}=p^0_i + 2 \cdot C_i $
        \ENDFOR
        \STATE \textbf{Return} $\mathcal{\hat{P}}=\{\hat{p_{i}}, i \in 1,2,...,N\}$ - prediction of the relaxed state position, 
\end{algorithmic}
\end{algorithm}

In order to reduce the edge representation ambiguity caused by linear aggregated node predictions and improve model robustness under graph structural variations, better model supervision and inference strategy are desired. In this work, with the prerequisite that all processes should be E(3)-equivariant, we propose to address the problem by designing a function $\phi$ that constructs a unique set of edge-wise label $\{\Delta p^\star_{e_{ij}} | e_{ij}\in\mathcal{E}\}$from node label$\{\Delta p^\star_{i}|i\in N\}$, to supervise the model in a finer-grained way. Reversely, the model shall generate an set of edge-wise prediction$\{\Delta \hat{p_{e_{ij}}} | e_{ij}\in\mathcal{E}\}$ that can reconstruct the node-wise prediction$\{\Delta \hat{p_{i}}|i\in N\}$. In this way, we are able to provide a concise label for each edge that is identical under different graph representations. At the same time, suppose there exists an oracle model $\mathbf{f}^\star$ such that as long as it can provide the optimal edge predictions $\hat{\Delta{p_{e_{ij}}}} = \Delta p_{e_{ij}}^\star$, we can always have $ \hat{\Delta p_i }=\Delta p_i^\star $ under different graph structure. We summarize three principles to follow for implementing such a training strategy: reversibility, uniqueness, and E(3)-equivariance.



\textbf{Reversibility:} After deconstructing the label of node $i$,$ \{\Delta p^\star_{e_{ij}}:j\in\mathcal{N}_i\}=\phi(\Delta p^\star_i)$, there needs to exist a function $\phi^{-1}$, s. t. $\Delta p^\star_i = \phi^{-1}(\{\Delta p^\star_{e_{ij}}:j\in\mathcal{N}_i\})$

\textbf{Uniqueness:} When performing label decomposition, for each $e_{ij}$, $\Delta p^\star_{e_{ij}}$ is identical under different graph representation $\mathcal{G}$. Moreover, for each $\{e_{ij}:j\in\mathcal{N}_i\}$, there does not exist another set of $\{\tilde{\Delta p^\star_{e_{ij}}}: j\in\mathcal{N}_i\}$ such that there exist $k \in \mathcal{N}_i$ and $\Delta p^\star_{e_{ik}} \neq \tilde{\Delta p^\star_{e_{ik}}}$ while $\phi^{-1}(\{\tilde{\Delta p^\star_{e_{ij}}}\})=\phi^{-1}(\{\Delta p^\star_{e_{ij}}\})=\Delta p^\star_i $

\textbf{E(3)-equivariance:} There exists a pair of group operations on nodes $\psi_v$ and edges $\psi_e$, such that when an E(3) transformation $\psi$ is performed to the N-body system, for both the deconstruction process and the reconstruction process,  we have $\phi(\psi_v(\Delta p^\star_i)) = \psi_e(\phi(\Delta p^\star_i))$ and 
$\psi_v(\phi^{-1}(\{\Delta p^\star_{e_{ij}}:j\in\mathcal{N}_i\})) = \phi^{-1}(\psi_e(\{\Delta p^\star_{e_{ij}}:j\in\mathcal{N}_i\}))$.

\subsection{Project and Fit}
\label{sec_proj_and_fit}
Based on the proposed principles that we discussed above, as illustrated in Figure \ref{pnf idea}, we instantiate an example schema, namely project and fit. The method performs label deconstruction from nodes to edges and reconstructs predictions from edges to nodes. The specific algorithm is summarized in Algorithm \ref{alg:proj} and \ref{alg:fit}.

For the label deconstruction part \ref{alg:proj}, we project the positional shift vector $\Delta p^\star_i$ of node $v_i$ on each outward edge direction with $\{\Delta p^\star_{e_{ij}}\} = \phi(\Delta p^\star_i)$ , and use the projection magnitude $\{\mathbf{m}^\star_{ij}=\Delta p^\star_i \cdot d_{e_{ij}} \in \mathbb{R} \}$as the label for edges, forming the projection magnitude matrix $\mathcal{M}^\star \in \mathbb{R}^{N\times N}$. Note that $\mathcal{M}^\star$ is not a symmetric matrix, and will be used in model supervision as the edge-wise label.


 For the label reconstruction part \ref{alg:fit}, the prediction of positional displacement projection on edges will be calculated by $\{\Delta \hat{p_{e_{ij}}} =\mathbf{\hat{m}}_{ij} \cdot d_{e_{ij}}  \in \mathbb{R}^3\}$, where $\mathbf{\hat{m}}_{ij}$ being the projection magnitude prediction. For each node $v_i$ we use a 3D sphere fitting algorithm to fit a sphere that is passing the original point $p_i^0$ while having the lowest $L^2$ distance with each endpoint of projection vectors. Note that the sphere is unambiguously defined by sphere center $C_i$. In this way, as a reverse process of the label deconstruction process, we derive $\hat{p}_i = p^0_i + 2 * C_i$. Compared with the linear aggregated node prediction shown in equation \ref{equation3}, our method proposes a non-linear node displacement prediction algorithm as
\begin{equation}
\hat{p}_i ={p}_i^0+
SphereFitting(\{ \mathbf{\overrightarrow{m}}_{ij} | j\in\mathcal{N}_j\}).\\
\end{equation}



In the project and fit algorithm, E(3) equivariance is naturally preserved for both the label decomposition part and the label recovery part since we are using the E(3)-equivariant value $d_{e_{ij}}$ as reference. Project and fit also fulfill the requirement that it generates a unique edge prediction for each direction. Finally, the label projection operation and sphere fitting operation are inverse to each other. For these reasons, as Figure \ref{pnf idea} (f) and (g) showed, compared with sum-aggregated node displacement prediction, a model trained with the DR-Label strategy can provide more robust results under the variation of the graph structure. Experimental evidence of our statements is further shown in section \ref{robustness of D&R}.

\section{Model Implementations}
\label{sec_model_impl}
\begin{figure*}[th]
\begin{center}
\centerline{\includegraphics[width=\textwidth]{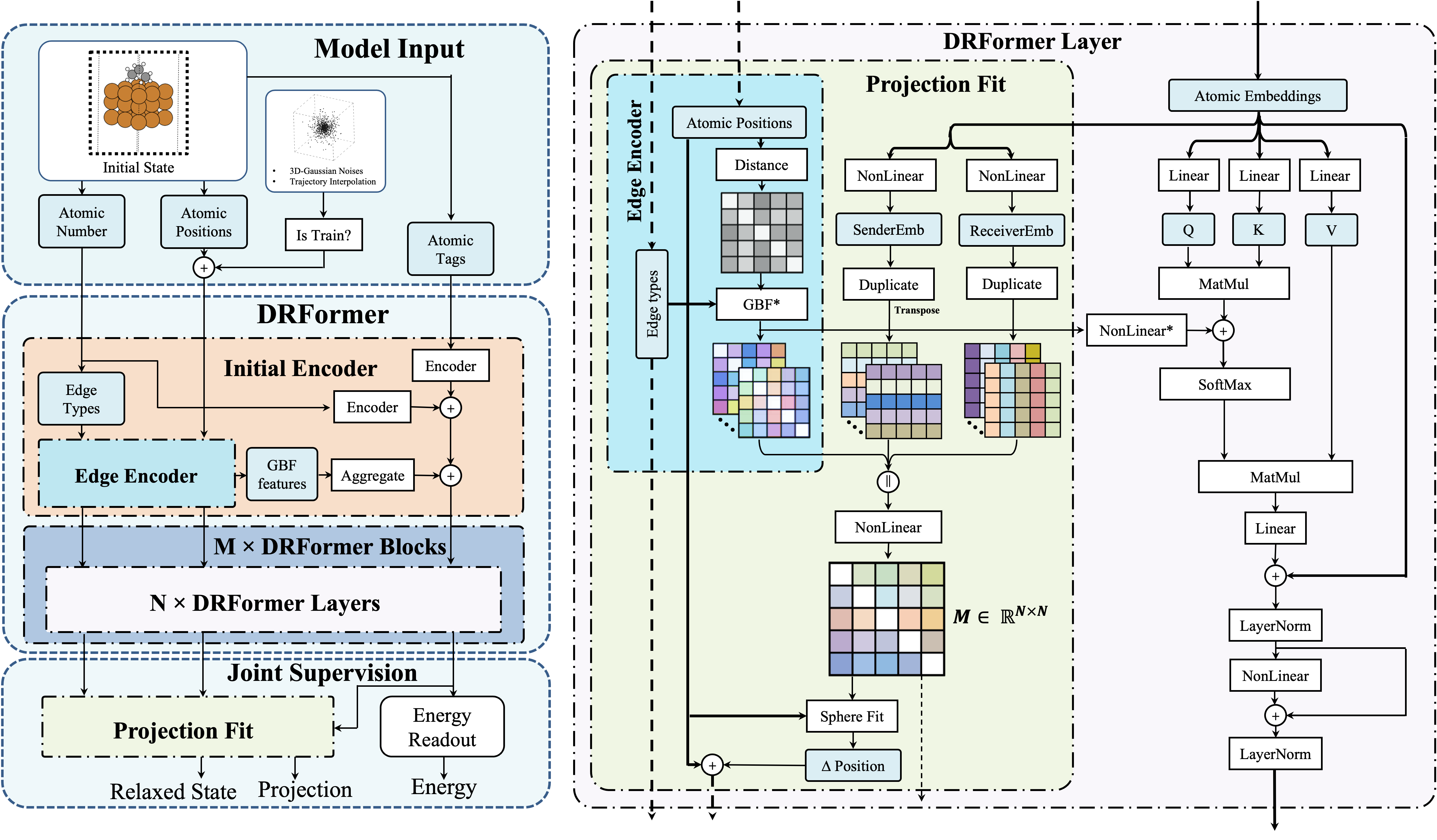}}
  \caption{Left: the overall architecture of DRFormer. Right: specific decomposition of the DRFormer Layer, edge encoder module, and the projection fit module. $||$ indicates concatenation and $+$ indicates addition. $\star$ indicates that the parameter of the corresponding part is shared across different layers in the model. }
  \label{DRFormer Layer}
\end{center}
\end{figure*}

To further examine the effectiveness of the DR-Label schema, we integrate it into three drastically different kinds of architectures that achieve outstanding performance: \textbf{Graphromer}, \textbf{GemNet}, and \textbf{SCN}. Based on DR-Label integrated graphormer, we further propose \textbf{DRFormer}, which attains state-of-the-art performance on both OC20 and SAA datasets. 

\subsection{DR-Label Integrated Models}

We use \textbf{Graphormer} as an example to illustrate our DR-Label integration, where the specific architecture is shown in the projection fit block of Figure \ref{DRFormer Layer}. The input of the block consists of three parts: the initial atomic positions $\mathcal{P}^0$, the edge types information generated based on the atomic numbers of the two endpoints, and the atomic embeddings for the graphormer. 

In the edge encoder block, atomic positions will be used to calculate a symmetric edge-level Euclidean distance matrix, where the $ij$ element indicates the distance between the $i$ atom and the $j$ atom. Next, the distance matrix will combine with the edge type matrix, and generate a gaussian basis function (GBF) based edge-level embedding, which contains the distance and endpoint type information of the edge. 

Note that in our project and fit method, decomposed label $\Delta p^\star_{e_{ij}}$ does not necessarily equal to $\Delta p^\star_{e_{ji}}$ because they are decomposed from the positional displacement of different nodes ($v_i$ and $v_j$). Therefore, the embedding matrix for edge prediction should not be symmetric. To stress this, we transform the final atomic embedding by two distinct nonlinear layers and generate node-level embedding for the sender node and receiver node of each edge, respectively. Finally, we generate the edge-level embedding in the order of $\{$GBF edge embedding $|$ sender node embedding $|$ receiver node embedding$\}$  by concatenation. The result will be fed into another non-linear layer to generate the single-valued asymmetric edge-wise magnitude matrix $\hat{\mathcal{M}} \in \mathbb{R}^{N\times N}$. Each element $\hat{m}_{ij}$ indicates the magnitude of $\Delta p^\star_i$ project in the direction of $d_{e_{ij}}$. Finally, the edge direction tensor will be combined with the edge-wise magnitude matrix into the sphere-fitting algorithm, and generate an E(3)-equivariant vector prediction $\Delta \hat{p}_i$ for each node. 

In the label decomposition part, another matrix $\mathcal{M}^\star\in\mathbb{R}^{N\times N}$ will be generated by the label projection algorithm as the supervision signal for the edge-wise magnitude matrix $\hat{\mathcal{M}}$. The node displacement prediction $\Delta \hat{\mathcal{P}}$ will be supervised by $\Delta \mathcal{P}^\star$. We use the original Graphormer implementation of the energy readout function to make graph-level supervision. Finally, we joint all three levels of model supervision by

\begin{equation}
\begin{gathered}
\mathcal{L}=\mathcal{L}_{G}+\lambda\mathcal{L}_{V}+\gamma\mathcal{L}_{E},
\end{gathered}
\end{equation}

where $\mathcal{L}$ refers to the total loss, and $\mathcal{L}_{G},\mathcal{L}_{V},\mathcal{L}_{E}$ refers to the graph-level, node-level, and edge-level loss respectively. $\lambda$ and $\gamma$ are used to control the relative intensity of different levels of supervision.

 Integrating DR-Label with GemNet-OC and SCN is generally the same as the above pipeline, where we keep most of the model unchanged. Integration of DR-Label strategy on GemNet-OC is straightforward. The model deployed stacks of GemNet interaction blocks, each outputs the embedding for each node and each edge, and finally aggregate representation across blocks with an MLP layer to generate final embedding for nodes and edges. We generate edge-wise predictions by $\mathbf{\hat{m}}_{ij}=
\mathbf{f}_{GemNet}([\mathbf{m}_{e_{ij}}||\mathbf{m}_{v_i}||\mathbf{m}_{v_j}])$, and follows the sphere fitting algorithm to generate $\hat{\mathcal{P}}$. Here $\mathbf{m}_{e_{ij}},\mathbf{m}_{v_i},\mathbf{m}_{v_j}$ stands for edge embeddings, sender node embedding, and target node embedding respectively. We further replaced the direct force supervision with our node-level displacement and edge-level projection supervision. 
 
 Integration DR-Label on SCN is slightly different from the above operation. SCN generates 128 evenly distributed directions on a unit sphere as $d_{e_{ij}}$, which we used as the outward edge directions for each node, and perform the label decomposition and reconstruction process. For each node $v_i$, the SCN model finally generates a $\mathbb{R}^{(L+1)^2 \times C}$ matrix representation of $C$ channels of functions on the sphere. The representation is first transformed by $\mathbf{f}_{SH2D}: \mathbb{R}^{(L+1)^2 \times C} \rightarrow  \mathbb{R}^{128 \times C}$ to transform the spherical harmonics to the representation over 128 discrete directions, changing the continuous function on sphere surface to 128 discrete directions. Then follows $\mathbf{f}_{D2M}: \mathbb{R}^{128 \times C}\rightarrow  \mathbb{R}^{128 \times 1}$ that transforms the $\textbf{m}_{ij}\in\mathbb{R}^{C}$ on the direction of $d_{e_{ij}}$ to the magnitude prediction $\hat{\textbf{m}}_{ij},j\in[1,128]$ . Again, we replaced the direct force supervision head with node-level displacement and edge-level projection supervision.

\subsection{The DRFormer Model}

\textbf{Intermediate Position Update}
Taking advantage of the graph structural variation robustness property of the DR-Label, we further develop the DRFormer model, which explicitly encodes and updates the atom coordinate along the model. The model structure is illustrated in Figure \ref{DRFormer Layer}. For each DRFormer layer, we utilize the projection fit block to update the atomic position on every $F$ layer of the DRFormer Layer. After the atomic position is updated, the GBF edge embedding will be changed accordingly for the graph attention part. Note that the blocks with $\star$ sign indicate that the parameter is shared across different layers. In this way, we generate a sequence of evolving intermediate position predictions, compelling our model to explicitly learn a physical interpretable trace along the model. 

\textbf{Noisy Node}
Following \cite{noisynodes}, we further add noise-augmented instances along each batch. The noise is only added in the training process. For atoms $v_i$ that have non-zero positional displacement label $\Delta p^\star_i$, we first perform a random interpolation between $p^0_i$ and $p^*_i$, then add a random 3D-gaussian noise $\sigma_i$ towards that position as a noisy initial state $\tilde{p^0_i}=p^0_i+\alpha\Delta p^\star_i+\sigma_i, \alpha \in [0,1]$. For these instances, we use the displacement between the noisy initial state and the target state $p^\star_0 - \tilde{p^0_i}$ as the node-wise supervision signal $\Delta \tilde{p^\star_i}$.



\section{Experiments and Results}
\label{sec_experiment}
\begin{table*}[th]
\caption{Result on OC20 IS2RE validation set}
\label{oc20_validation}
\begin{center}
  \begin{adjustbox}{width=\textwidth,center}
\begin{small}
\begin{sc}
\begin{tabular}{l|ccccc|ccccc}
\toprule
                 & \multicolumn{5}{c|}{MAE (eV) $\downarrow$}               & \multicolumn{5}{c}{AEwT (\%) $\uparrow$}               \\ \cmidrule{2-11} 
Model            & ID     & OOD Ads. & OOD Cat. & OOD Both & Average & ID   & OOD Ads. & OOD Cat. & OOD Both & Average \\
\midrule
Graphormer & $ 0.4329 $ & $ 0.5850 $   & $ 0.4441 $   & $ 0.5299 $   & $ 0.4980 $  & $ - $ & $ - $     & $ - $     & $ - $     & $ - $         \\
Graphormer+DR-Label & $ 0.4574$       & $ 0.5497$          & $0.4686 $         & $0.4873 $          & $ 0.4907$        & $ 6.34 $     &$ 3.95$          &$ 6.02$          &$ 4.34$          & $ 5.13 $        \\
\midrule
GNS & $ 0.54 $  & $ 0.65 $    & $ 0.55 $    & $ 0.59 $    & $ 0.5825 $   & $ - $ & $ - $     & $ - $     & $ - $     & $ - $        \\
GNS+Noisy Nodes & $ 0.47 $  & $ 0.51 $    & $ 0.48 $    & $ 0.46 $    & $ 0.4800 $   & $ - $ & $ - $     & $ - $     & $ - $     & $ - $         \\
Equiformer & $ 0.4222 $ & $ 0.5420 $   & $ 0.4231 $   & $ 0.4754 $   & $ 0.4657 $  & $ 7.23 $ & $ 3.77 $     & $ 7.13 $     & $ 4.10 $     & $ 5.56 $         \\
Equiformer+Noisy Nodes & $ 0.4156 $ & $ \underline{0.4976} $   & $ \textbf{0.4165} $   & $ \underline{0.4344} $   & $ \textbf{0.4410} $   & $ 7.47 $ & $ 4.64 $      & $ 7.19 $     & $ 4.84 $     & $ 6.04 $        \\
\midrule
DRFormer w/o InterPos & $ \textbf{0.4085}$       & $ 0.5149$          & $\underline{0.4286} $         & $0.4635 $          & $ 0.4539$        & $ \textbf{9.08} $     &$ 4.87$          &$ \textbf{8.36}$          &$ \underline{5.11}$          & $ \textbf{6.86} $        \\
DRFormer & $ \underline{0.4187}$       & $ \textbf{0.4863}$          & $0.4321 $         & $\textbf{0.4332} $          & $ \underline{0.4425}$        & $ \underline{8.39} $     &$ \textbf{5.42}$          &$ \underline{8.12}$          &$ \textbf{5.44}$          & $ \underline{6.84} $        \\

\bottomrule
\end{tabular}
\end{sc}
\end{small}
\end{adjustbox}
\end{center}
\end{table*}

\begin{table*}[th!]
\caption{Result on OC20 IS2RE testing set}
\label{oc20_test}
\begin{center}
  \begin{adjustbox}{width=\textwidth,center}
\begin{small}
\begin{sc}
\begin{tabular}{l|ccccc|ccccc}
\toprule
                 & \multicolumn{5}{c|}{MAE (eV) $\downarrow$}               & \multicolumn{5}{c}{AEwT (\%) $\uparrow$}               \\ \cmidrule{2-11} 
Model            & ID     & OOD Ads. & OOD Cat. & OOD Both & Average & ID   & OOD Ads. & OOD Cat. & OOD Both & Average \\

\midrule
GemNet-OC & $ 0.560 $  & $ 0.711 $    & $ 0.576 $    & $ 0.671 $    & $ 0.630 $   & $ 4.15 $ & $ 2.29 $     & $ 3.85 $     & $ 2.28 $     &  $3.14$        \\
GemNet-OC + DR-Label & $0.450$       & $0.707$         & $0.478$         & $0.638$        &   $0.568$      & $6.55$      & $2.56$         &  $5.93$        & $ 2.98$         & $4.51$        \\
SCN & $ 0.516 $  & $ 0.643 $    & $ 0.530 $    & $ 0.604 $    & $ 0.573 $   & $ 4.92 $ & $ 2.71 $     & $ 4.42 $     & $ 2.76 $     &   $3.70$         \\
SCN + DR-Label & $0.474$       & $0.675$          & $0.482$         & $0.626$          & $ 0.564 $        & $5.55$    & $2.63$         &  $5.13$        & $2.69$         &  $ 4.00 $        \\
\midrule
Graphormer-3d (Ensemble) & $ 0.3976$       & $ 0.5719$          & $0.4166$         & $ 0.5029$          & $0.4722$        & $8.97$    &  $3.45$        &  $8.18$        & $3.79$         & $6.1$         \\
GNS+Noisy Nodes & $0.4219 $       & $ 0.5678 $          & $ 0.4366$         & $\textbf{0.4651} $          & $ 0.4728$        & $\underline{9.12}$    & $\textbf{4.25}$         &  $8.01$        & $\textbf{4.64}$         & $\textbf{6.5}$        \\
Equiformer+Noisy Nodes & $0.4171 $       & $0.5479 $          & $0.4248 $         & $0.4741 $          & $ 0.4660$        & $7.71$    & $3.70$         & $7.15$         & $4.07$         & $5.66$        \\
\midrule
DRFormer w/o InterPos & $ \underline{0.3928}$       & $ \underline{0.5523}$          & $\underline{0.4113} $         & $0.4766 $          & $ \underline{0.4582}$        & $ 8.77 $     &$ 3.81$          &$ \underline{8.32}$          &$ 3.97$          & $6.21$        \\
DRFormer & $ 0.4148 $       & $ 0.5691$          & $0.4365 $         & $0.4931 $          & $ 0.4783$        & $ 8.07 $     &$ \underline{4.09}$          &$ 7.31$          &$ 4.30$          & $ 5.94$        \\
DRFormer-Average & $ \textbf{0.3865}$       & $ \textbf{0.5435}$          & $\textbf{0.4060} $         &$\underline{0.4677} $          & $ \textbf{0.4509}$        & $ \textbf{9.18} $     &$ 4.01$          &$ \textbf{8.39}$          &$ \underline{4.33}$          & $\underline{6.48}$        \\
\bottomrule
\end{tabular}
\end{sc}
\end{small}
\end{adjustbox}
\end{center}
\end{table*}

\subsection{Experimental Settings}

\subsubsection{Dataset}
Our method and models are mainly evaluated over the catalyst adsorption energy prediction problem. Two specific datasets are used to examine our proposed methods. First is the publicly available OC20 dataset\cite{chanussot2021OC20}, on which we conduct our main experiments. The dataset contains 460k instances for IS2RE methods training and has 4 different kinds of datasets for evaluation: the in-distribution dataset (ID) and three out-of-distribution datasets that have unseen structures in adsorbates (OOD-Ads), catalyst slabs (OOD-Cat), or both (OOD-Both). We followed the official validation and testing split to evaluate our model. Another dataset we used is a smaller dataset from \cite{SAA_dataset} that predicts the CO adsorption energy of Cu-based single-atom alloy (SAA) catalysts. The SAA dataset is composed of 41 doping species on different sites of 5 surfaces of the ideal Cu crystal. CO molecules are adsorbed on different positions of the surfaces, leading to 3075 instances in total. Each sample includes a DFT calculated adsorption energy as the label, as well as corresponding initial (unrelaxed) and relaxed configurations for training. The catalyst structures in OC20 dataset and the SAA dataset are fundamentally different, since SAA catalysts are not included in the former one, which only considers regular bulk catalyst surfaces.


\subsubsection{Model Evaluation Settings}
Following the settings of OC20, we use the mean absolute error of adsorption energy prediction(MAE, eV) and the percentage of the absolute error of the predicted energies lies within the threshold of 0.02eV (AEwT,\%) as our evaluation metric.

For the experiments on OC20 datasets, our testing set results have been reported based on our submission to the official evaluation server on \textit{eval.ai}\cite{yadav2019evalai}. For the experiments on the SAA datasets, we followed the experiment setting from \cite{mtmd-saa}  and made 10 random splits with train:validation: test = 60\%:20\%:20\%, report the mean and standard deviation of the result. 

\subsection{OC20 Results}
We report our result on OC20 validation and testing set in Table \ref{oc20_validation} and \ref{oc20_test}, respectively. Graphormer performance is only compared on the validation set since the official single-model testing set result is not reported. We use \textbf{bold} font for top-performed results and \underline{underline} for runner-up results. "+ DR-Label" indicates the DR-Label module is integrated with the corresponding model; "w/o InterPos" means the intermediate positional update is not included in the DRFormer model. Finally, for the testing set, we average the prediction of DRFormer with/without the intermediate positional update and report the result in DRFormer-Average.

The first thing we observe is that with the DR-Label module integrated, the overall performance was constantly improved for Graphormer, GemNet-OC, and SCN. More specifically, by applying the DR-Label module, we find that GemNet-OC gained a 57.8\% relative increase of AEwT value on the ID dataset and 54.0\% on the OOD-Cat dataset. The results reveal that DR-Label guides the model to learn the underlying physics of relaxation and helps the model to provide a more precise adsorption energy prediction. 

Also, we observe that the DRFormer achieves the new state-of-the-art performance of OC20 IS2RE validation set by a large margin on the AEwT metric. This indicates that our method can provide a much large portion of accurate adsorption energy prediction. We also notice that our result is either top-performed or very close to the top-performed Equiformer+Noisy Nodes, without equivariant features or attention added.
Finally, by averaging the predictions of two DRFormer models with/without the InterPos module, we reached a new state-of-the-art performance on the OC20 IS2RE testing set. More importantly, Graphormer-3D(Ensemble) used 31 separately trained models to provide the final result, which requires 46.5 days on an $8\times A100$ GPU machine. On the contrary, our models only require up to 5 days for training under the same setting.

\subsection{SAA Dataset Results}
We keep the model structure exactly the same as it was in the OC20 experiments and conduct experiments on the SAA dataset to further examine our method, reporting the results in Table \ref{SAA_test}. We benchmark the Graphormer performance on the SAA dataset and train the DRFormer model. We observe that Graphormer does not achieve top performance, while DRFormer reaches a new state-of-the-art performance on this dataset. In DRFormer-FT,  we load the final checkpoint trained on the OC20 dataset and fine-tune the pre-trained model on the SAA dataset. We find that the final performance is improved by a large margin, reaching a 20.2\% decrease of MAE and a 16.2\% increase of AEwT compared with the previous best method MT-MD DimeNet++.

\begin{table}[t]
\caption{SAA dataset performance}
\label{SAA_test}
\begin{center}
  \begin{adjustbox}{width=\columnwidth,center}
\begin{small}
\begin{sc}
\begin{tabular}{l|c|c}
\toprule
 Model                & {MAE (eV) $\downarrow$}    & {AEwT (\%) $\uparrow$}               \\ 

\midrule
Localized-cos + GBR & $ 0.120 \pm 0.018 $  & $ 29.9\pm1.8   $          \\
Gaussian-cos + GBR & $ 0.130 \pm 0.019 $  & $ 27.1\pm2.1   $          \\
Gaussian-tanh + GBR & $ 0.132 \pm 0.020 $  & $ 26.8\pm1.8   $          \\
SchNet & $ 0.257 \pm 0.113 $  & $ 10.3\pm7.7   $          \\
CGCNN & $ 0.182 \pm 0.089 $  & $ 14.7\pm3.7   $          \\
DimeNet & $ 0.099 \pm 0.016 $  & $ 32.9\pm2.8   $          \\
DimeNet++ & $ 0.094  \pm 0.015 $  & $ 33.7\pm1.5   $          \\
MT-MD DimeNet++ & $ 0.087 \pm 0.011 $  & $ 39.8\pm2.5   $          \\
\midrule
Graphormer & $ 0.0984  \pm 0.0156 $  & $ 33.0 \pm 1.8 $          \\


DRFormer & $ \mathbf{0.0828  \pm 0.0138} $  & $ \mathbf{44.5 \pm 2.1} $          \\
\midrule
DRFormer-FT & $ \mathbf{0.0694  \pm 0.0154} $  & $ \mathbf{56.0 \pm 2.3} $          \\
\bottomrule
\end{tabular}
\end{sc}
\end{small}
\end{adjustbox}
\end{center}
\end{table}

\subsection{Ablation Studies}
To further examine the effective module in the model, we perform an ablation study by training on the OC20 IS2RE 10k dataset and evaluate the OC20 IS2RE validation set. 

We observe that compared with the model without the DR-Label module, models with the DR-Label module have constantly improved performances. The intermediate node position update operation also provides an improvement in model performance, especially on ID and OOD-Cat datasets. Finally, we note that the noisy-node based data augmentation method provides a remarkable improvement in model performance.


\begin{table}[t]
\caption{Ablation Study}
\label{Ablation_table}
\begin{center}
  \begin{adjustbox}{width=\columnwidth,center}
\begin{small}
\begin{sc}
\begin{tabular}{cl|ccccc}
\toprule
\multicolumn{2}{c|}{Module Name}    & \multicolumn{5}{c}{Model}                           \\ \midrule
\multicolumn{2}{c|}{DR-Label}     & $\times$ & $\surd$ & $\times$ & $\surd$ & $\surd$ \\ 

\multicolumn{2}{c|}{Noisy Nodes}   & $\times$ & $\times$ & $\surd$ & $\surd$ & $\surd$ \\ 

\multicolumn{2}{c|}{Intermediate Pos}  & $\times$ & $\times$ & $\times$ & $\times$ & $\surd$ \\ 
\midrule
\multicolumn{1}{c|}{\multirow{5}{*}{MAE (eV)$\downarrow$}} & ID        &   $0.693$       & $ 0.688$          & $0.673 $         &  $ 0.664$    &$\textbf{0.648}$         \\
\multicolumn{1}{c|}{}                               & OOD Ads.         & $0.753 $         & $ 0.733$         &  $ 0.722 $        & $ 0.705$     &$\textbf{0.694}$         \\
\multicolumn{1}{c|}{}                               & OOD Cat.         &  $0.665$         &  $ 0.678 $        &  $ 0.652$        & $ 0.646$     &$\textbf{0.631}$        \\
\multicolumn{1}{c|}{}                               & OOD Both         & $0.669$         & $ 0.656$         & $ 0.651$         & $ 0.635$       &$\textbf{0.627}$      \\
\multicolumn{1}{c|}{}                               & Average          & $0.695$          & $ 0.689$         &  $ 0.675$        & $ 0.662$         &$\textbf{0.650}$         \\ \bottomrule
\end{tabular}
\end{sc}
\end{small}
\end{adjustbox}
\end{center}
\end{table}

\begin{figure}[t]
\begin{center}
\includegraphics[width=\columnwidth]{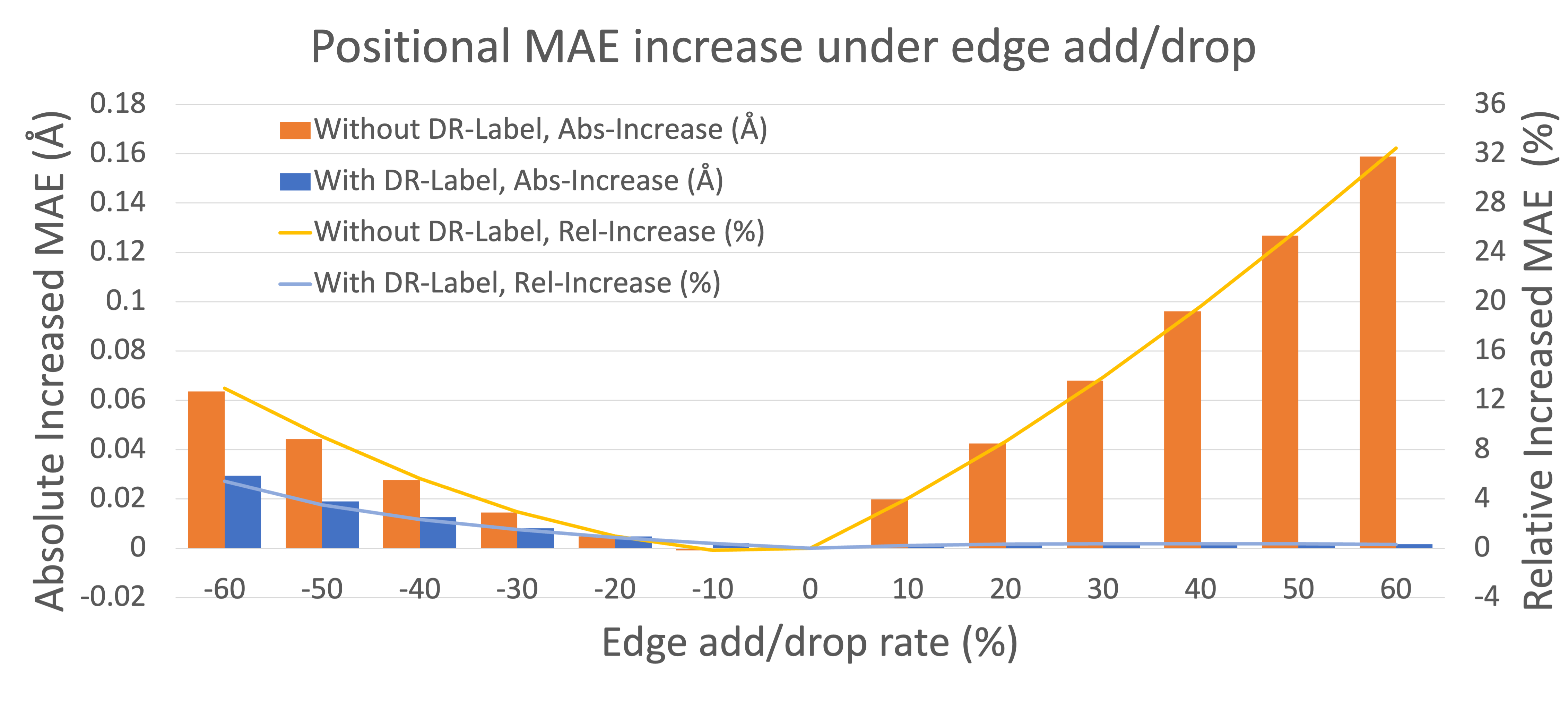}
  \caption{Position MAE increase under edge add/drop. DR-Label drastically slows down the rate of MAE increase.
  }
  \label{edge removal fig}
  
\end{center}
\end{figure}

\subsection{Robustness Induced by DR-Label}
\label{robustness of D&R}
From the discussion in section \ref{method}, we state that models would be more robust under graph structural variations by applying DR-Label. In this experiment, we study the behavior of models with and without the DR-Label module under random edge addition/removal and graph statistics variation by attaching a boolean matrix $B\in\{1,0\}^{N\times N}$ with each instance. 

Graphormer constructs the graph in a fully-connected way. Therefore, as a simulation of edge addition, we add additional $e_{ij-add}$ between $v_i$ and $v_j$, transforming the original graph into a multi-graph. Consequently, for model without DR-Label module, the attention weight $\alpha_{e_{ij}}$ will be multiplied by 2 whenever $B_{e_{ij}}=1$. For model with DR-Label module, edge addition corresponds to duplicate another $\hat{p}_{e_{ij}-add}=\hat{p}_{e_{ij}}$ when $B_{e_{ij}}=1$ before the sphere fitting operation. Similarly, when performing edge removal, for models without DR-Label, the attention weight $\alpha_{e_{ij}}$ will be set to zero. For models with the DR-Label module, $\hat{p}_{e_{ij}}$ will be set to zero. We denote the original graph structure as $\mathcal{G}$ and the edge-modified graph as $\tilde{\mathcal{G}}$, restore the final checkpoint of the Graphormer model and the Graphormer+DR-Label model trained on the OC20-10k dataset. We compare $\mathcal{L}_v$(\AA), the $L^2$-MAE of node position prediction on the ID validation dataset under edge removal, and summarize the result in Figure \ref{edge removal fig}. The left axis corresponds to the absolute node $L^2$-MAE increase $\Delta\mathcal{L}_v=\mathcal{L}^{\tilde{\mathcal{G}}}_v-\mathcal{L}^\mathcal{G}_v$ under edge add/drop. The right axis refers to the relative increase of node $L^2$-MAE towards the original graph structure($\Delta\mathcal{L}_v/\mathcal{L}^\mathcal{G}_v$). 

The figure shows that DR-Label drastically refrained the node predictions from collapse under edge addition. For edge dropping, we found that the DR-Label module slows down the increase of node-wise MAE. When adding 60\% edges, the absolute increase of MAE for the model without DR-Label is $9452\%$ larger than the model with DR-Label.  The result is in accord with our argument in \ref{method} and Figure \ref{pnf idea}(f) and (g) that DR-Label tends to learn a model that is robust under graph structural variation.

\section{Conclusions}
In this work, we introduce the DR-Label strategy, an effective module for improved equilibrium state prediction models of catalysis systems. DR-Label help alleviates the problem of non-uniqueness of edge representation and improves model robustness under graph structural variations. Three different models benefit significantly from integrating the module. We further propose the architecture of DRFormer, which achieved a new state-of-the-art performance of adsorption energy prediction on both the OC20 dataset and the SAA dataset. We expect DR-Label could be used as a general module when building GNN models for equilibrium state prediction problems for atomic systems, and potentially further promote the research of edge-level auxiliary supervisions of general GNN models.



\nocite{langley00}

\bibliography{example_paper}
\bibliographystyle{icml2022}

\end{document}